\pdfoutput=1
\documentclass[11pt]{article}

\usepackage{acl}

\usepackage{times}
\usepackage{latexsym}

\usepackage[T1]{fontenc}
\usepackage[utf8]{inputenc}

\usepackage{microtype}

\title{Moving beyond word lists: towards abstractive topic labels for human-like topics of scientific documents}

\author{Domenic Rosati \\
  scite.ai / Brooklyn, NY}

\begin{document}
\maketitle
\begin{abstract}
Topic models represent groups of documents as a list of words (the topic labels). This work asks whether an alternative approach to topic labeling can be developed that is closer to a natural language description of a topic than a word list. To this end, we present an approach to generating human-like topic labels using abstractive multi-document summarization (MDS). We investigate our approach with an exploratory case study. We model topics in citation sentences in order to understand what further research needs to be done to fully operationalize MDS for topic labeling. Our case study shows that in addition to more human-like topics there are additional advantages to evaluation by using clustering and summarization measures instead of topic model measures. However, we find that there are several developments needed before we can design a well-powered study to evaluate MDS for topic modeling fully. Namely, improving cluster cohesion, improving the factuality and faithfulness of MDS, and increasing the number of documents that might be supported by MDS. We present a number of ideas on how these can be tackled and conclude with some thoughts on how topic modeling can also be used to improve MDS in general.
\end{abstract}

\section{Introduction}
Topic modeling, a common approach for extracting themes from scientific documents, is currently facing many challenges: methodological validity \citep{shadrova_topic_2021}, validity of automated evaluation \cite{doogan_topic_2021, hoyle_is_2021}, and utility of classical approaches \citep{sia_tired_2020, zhang_is_2022}. We propose an additional challenge: \textit{are lists of words the best we can do for topic labels?}

Topic models have tended to represent a topic as a list of words. Traditional topic labels are supposed to be “a set of terms, when viewed together, enable human recognition of an identifiable category” \citep{hoyle_is_2021}. However, a set of terms do not align with our intuitive understandings of what a topic is: a common theme or concept explicated as a word, phrase, or natural language description \citep{shadrova_topic_2021}. In this paper, we present an exploratory case study using multi-document summaries (MDS) as labels for clusters of citations in order to understand current limitations and future work needed for using abstractive topic labels for human-like topics of scientific documents. To our knowledge, it is the first work that proposes to use MDS for topic labeling on top of topic clusters constructed with contextualized embeddings.

In addition to word lists not aligning with natural understanding of what a topic is, \citet{shadrova_topic_2021} has presented an extensive criticism of why traditional topic models based on lexical overlap measures lead to problematic topic models. Namely that they they \textit{fail to understand word sense and capture context}. Recent approaches have relaxed these restrictions when constructing topic clusters \citep{bianchi_pre-training_2021,grootendorst_bertopic_2022} by using contextualized word embeddings. However topic labels in those models are still constructed as word lists drawn from documents such as through TF-IDF.

Some work has anticipated this challenge by developing topic representations with phrases \citep{popa_bart-tl_2021} and summaries \citep{basave_automatic_2014, gourru_united_2018, wan_automatic_2016}. But those works tend to be extractive, drawing the phrase or summary from a single document in the cluster\footnote{see \citet{alokaili_automatic_2020,popa_bart-tl_2021} for recent abstractive works.}. In the extractive setting, \textit{there may be no existing and fluent phrase or sentence that is capable of describing all documents in the cluster} or there may be multiple and even conflicting subtopics in the cluster that require a longer abstractive representation for producing a factual summary.

\section{Proposed Method}
\label{method}

\subsection{Topic modeling as clustering and MDS}

In order to address the issues presented above, we propose using abstractive MDS as an approach to topic labeling. Topic modeling can be reframed as a set of two tasks: (1) finding meaningful clusters for documents \citep{sia_tired_2020, zhang_is_2022} and (2) performing MDS on those individual clusters to find meaningful topic labels. In this framework, LDA \cite{blei_latent_2003} uses document-word distributions to construct clusters and word lists drawn from those clusters as a form of MDS. Since we are looking at abstractive MDS that moves beyond word lists, we propose that the \textit{topic representation be a sentence or paragraph} but there is no reason why an abstractive MDS can't be trained to generate phrases or even word lists (see \citet{alokaili_automatic_2020}) since word lists may still be appropriate in some situations.

In order to accomplish this, one can first use an approach for document clustering that uses contextualized word embeddings to avoid the issues mentioned above. By separating the clustering step from the representation step, we can use separate measures of cluster coherence to evaluate the quality of document clusters before we proceed to topic representation. We can also use evaluations of resulting topic representations later as an additional step to inspect the quality of our topic clusters.

After obtaining document clusters, MDS models such as \citep{lu_multi-xscience_2020} can be used to produce natural language summaries that synthesize common themes from documents. Recent work on MDS within the scientific and biomedical domain \citep{deyoung_ms2_2021, lu_multi-xscience_2020, shen_multi-lexsum_2022} show good results in producing both single sentence (extreme) summaries as well as long form summaries over many scientific documents.

\subsection{Evaluation}

Topic model evaluation is challenging (see \citet{chang_reading_2009, hoyle_is_2021, doogan_topic_2021}). Traditional metrics like coherence (NPMI), perplexity, and diversity scores are studied in the context of topic word lists and validated with correlation to human ratings of the utility or coherence of those topic word lists. Since we suggest developing abstractive topic representations, we want a way to compare various forms of both abstractive and extractive topic representations presented by the model. Since we are treating representation as a summarization task and this task includes measures that work across extractive and abstractive settings, we suggest that we start with standard summarization metrics such as overlap metrics like Rogue (as used in \citet{cui_topic-guided_2021} or semantic metrics such as BERTScore \citep{zhang_is_2022} (as used in \citet{alokaili_automatic_2020}).

\section{Case study: how has a scientific document been cited?}

To evaluate our proposed method, we chose topic modeling over scientific documents as a setting. While several methods exist for determining citation intent function \citep{basuki_sdcf_2022, nicholson_scite_2021} and the relationship between two papers \citep{luu_explaining_2021}, there is very little work on topic models over citations (for some representative work on "citation summary" see \citet{elkiss_blind_2008, wang_generating_2021, zou_citation_2021}). Topic representations of citations are interesting for characterizing trends in how a paper has been cited or helping researchers identify relevant citations to read among potentially thousands of other citations. In this work, we treat topic labels as a "citation intent" label and use the proposed approach to understand the utility of MDS for topic modeling in this setting.

\section{Experimental Setup}

\begin{table}[]
\begin{tabular}{lll}
\hline
Model & Source \\ \hline
multi-lexsum-long & \citet{shen_multi-lexsum_2022} \\
multi-lexsum-tiny & \citet{shen_multi-lexsum_2022} \\
ms2 & \citet{deyoung_ms2_2021} \\
multixscience & \citet{lu_multi-xscience_2020} \\
topic lists & \citet{bianchi_pre-training_2021} \\ \hline
\end{tabular}
\caption{
\label{models_table}
Generative models used for abstractive MDS topic representations.
}
\end{table}

We apply the method described in section \ref{method} in order to identify clusters of citations and provide labels for those clusters without any supervision. Specifically, we present a case study of what this looks like on a single paper to illustrate the potential of our approach and try to assess future work needed in order to make MDS a good solution for topic labeling in general and citation summarization in particular.

For this study, we used scite.ai \citep{nicholson_scite_2021} to extract in-text passages which contained citations (citation statements) to the paper \cite{lau_machine_2014}, a well known paper that introduces the NPMI metric in topic modeling. This resulted in 183 citation statements which is the corpus we will use for topic modeling.

In order to identify meaningful groups of clusters we use contextualized topic models (CTM) \citep{bianchi_pre-training_2021} since this method uses contextualized word embeddings (we used SPECTER for constructing embeddings \citep{cohan-etal-2020-specter}). We selected CTM since we still get word lists as topic labels which we used for evaluation. In order to select the number of topics hyperparameter, we trained CTM several times steadily increasing the number of topics from 3 to 50 and selected the best model according to coherence (NPMI) resulting in a 10 topic model (see Appendix \ref{ctm} for more details) over 183 citation statements.

The models selected for generating abstractive MDS are outlined in Table \ref{models_table}. All MDS models used are based on the longformer architecture \citep{beltagy_longformer} and used beam search (5 beams) with greedy decoding.

\section{Results}

\begin{table}[]
\begin{tabular}{lll}
\hline
Model & R-1 & BERTScore \\ \hline
multi-lexsum-long & 38 & 85 \\
multi-lexsum-tiny & 3 & 81 \\
ms2 & 3 & 81 \\
multixscience & 15 & 80 \\
topic lists & 1 & 76 \\ \hline
\end{tabular}
\caption{
\label{study_evaluation}
Rouge-1 (R-1) and BERTScore (F1) results for each models topic representations measured against.
}
\end{table}

\begin{table*}[]
\centering
\begin{tabular}{p{0.99\linewidth}}
\hline
\textbf{Topic 0}\\
 NPMI and Topic Coherence are measures used to measure the semantic coherence of topics. \\
\textbf{Topic 4 }\\ 
Topic model quality and interpretability are two different metrics used to measure the semantic interpretability of a topic. \\
\textbf{Topic 2} \\ Evaluation metrics: Log predictive probability (LPP) and topic interpretability \\ \hline
\end{tabular}
\caption{
\label{topic_evaluation}
Topic representations produced by multi-lexsum-tiny. Compared to word lists they are much more readable and closer to everyday notions of topics.
}
\end{table*}

Table \ref{study_evaluation} shows the Rogue-1 (R-1) and BERTScore (average F1 across topics) for each of the models selected for generating topic representations using MDS as well as the topic lists generated by CTM. It is important to underscore that R-1 and BERTScore are not validated against human studies for topic representations and this is simply a small case study on what an approach might look like. In spite of this, our results paint an initial picture of how these methods perform, especially when compared to model outputs (see Appendix \ref{outputs} for samples). Topic word lists have the worst R-1 and BERTScore. The MDS models do a little bit better with multi-lexsum-long having the best overall score. multixscience also does well with regards to R-1. Since multixscience and multi-lexsum-long are long form summaries, it appears that R-1 is potentially biased towards longer summaries and may not be a good measure across representations, in particular it may be uninformative for evaluating the performance of topic lists. ms2 and multi-lexsum-tiny are smaller and have better BERTScore then multixscience indicating they might provide more semantically similar representations. We are also not sure whether BERTScore suffers from the same bias towards longer or more sentence-like inputs.

We randomly sampled 3 topics to explore their representations. As an example, table \ref{topic_evaluation} shows representations using the multi-lexsum-tiny model (full details are available in Appendix \ref{outputs}. In representations for topic 0 (Table \ref{topic_0}), we see there is a general agreement across models that the citing documents are discussing measurement. We can see that the topic representations appear to be split between measuring interpretability (multixscience, multi-lexsum-long) and those discussing the correlation between measures (ms2, multi-lexsum-long) or even potentially an additional topic of describing measures used (multi-lexsum-tiny). Conflicting summaries are not surprising given issues in MDS with regards to summarizing diverse and potentially conflicting documents \citep{deyoung_ms2_2021}. Table \ref{topic_0} shows a diversity of topic labels that might be appropriate under different scenarios of applying topic models. Labels like the ones in Table \ref{topic_evaluation} might be useful for labels that are easy and fast to read while longer summaries in multixscience and multi-lexsum-long might be useful for users who want to engage deeper.

\section{Discussion}
\label{discussion}

In order to ensure downstream topic labels are coherent, document clusters must represent meaningful and well separated clusters. \citet{grootendorst_bertopic_2022, sia_tired_2020, zhang_is_2022} have shown that traditional clustering methods might provide good candidates for moving beyond topic models like LDA that suffer from lack of contextualized natural language understanding due to their use of word co-occurence statistics for constructing topic clusters. However in order to fully replace traditional methods we would like to see: (1) the demonstration of effective mixed-membership approaches in abstractive topic modeling to recover the ability for documents to belong to multiple topic clusters, (2) the demonstration of cluster evaluation measures that correlate well with how humans might group documents and possibly (3) the development of fully learnable architectures where clustering might be learned with feedback from topic representation quality.

\citet{deyoung_ms2_2021} has shown that MDS struggles with factual consistency. We see an opportunity for topic clustering as a step before performing MDS as a potential method for improving factual consistency since a contradicting source document that would normally be in the document set might be separated out with initial topic clustering. Furthermore, initial topic clustering might provide a way for developing more granular multi-aspect summarization techniques by clustering documents by aspect. Either way, we are weary of the known issues with factuality in MDS (\citet{deyoung_ms2_2021}) especially in the scientific domain where factual consistency is critical. To develop our approach along these lines, we suggest continuing to extend evaluation of factuality and faithfulness to the MDS setting (as identified in \citep{deyoung_ms2_2021}).

In order to make this approach work for a wide variety of application and analysis scenarios, controllable summarization (such as \citet{keskar_ctrl_2019}) should be investigated so that users can control for length of summaries (such as question, phrase, sentence, or paragraph) or style of summary (such as in the style of a paper title, abstract, citation, or literature review). Additional controls such as the ones suggested in \citet{shadrova_topic_2021} like granularity of topic label can also be developed in a controllable summarization framework in such a way as to make topic representations better fit for user's needs.

Finally while methods like longformer \citep{beltagy_longformer} enable the use of transformers with multiple documents as input, more research needs to be done to enable a method like the one we proposed on large sets of documents. In the scientific domain, where we might want to model hundreds or even thousands of full-text articles belonging to a single cluster, the approaches presented would be intractable without further development of long-attention transformer models.

One advantage of our approach is that since we are breaking topic models out into clustering and MDS as separate steps we can rely on a established work for evaluation of document clusters and summaries to assess models performance. While we'd need to validate the application of these metrics in end-to-end topic modeling scenarios, if text clustering and summarization metrics do correlate with human judgements of topic cluster and representation quality then we can avoid using topic modeling metrics which have come into question repeatedly \citet{chang_reading_2009, hoyle_is_2021, doogan_topic_2021}). However, we will not know this until we design robust human studies to validate the approach we have proposed above.

\section{Conclusion}

In this paper, we presented a reframing of topic modeling as document clustering with MDS applied to produce topic representations that might (1) align more intuitively with what humans understand as topics and (2) overcome some of the issues with topic models using bag of word assumptions such as inability to capture context. An initial case study on using this approach for unsupervised discovery of citation intents was explored. We found that while cohesive alternatives to topic representations can be produced using MDS in a variety of styles (short and long summaries), there are still many obstacles that need to be overcome before we can fully evaluate whether this approach could provide a viable alternative to traditional topic modeling and representation. Namely, improving cluster cohesion, improving the factuality and faithfulness of MDS, and increasing the number of documents that might be supported by MDS. While there might be an advantage in utilizing well validated approaches for evaluating clustering and summarization as measures of our approach, future studies will need to validate those with human studies. It is our hope that further work in this area can use our discussion as a roadmap towards what needs to be done if we want to move past word lists as topic representations.

\bibliography{anthology,custom}
\bibliographystyle{acl_natbib}

\appendix

\section{Topic Model Selection}

\label{ctm}

\begin{table}[]
\begin{tabular}{lll}
\hline
coherence (NPMI) & diversity & topics \\ \hline
-0.25 & 0.97 & 10 \\
-0.27 & 0.97 & 20 \\
-0.32 & 0.96 & 25 \\
-0.32 & 0.97 & 15 \\
-0.35 & 1.00 & 5 \\
-0.38 & 1.00 & 3 \\
-0.38 & 0.97 & 50 \\ \hline
\end{tabular}
\caption{
\label{ctm_training}
Selecting CTM topic model by evaluating CTM coherence (NPMI) and diversity on different topic numbers parameters.
}
\end{table}

Table \ref{ctm_training} describes the evaluation of all the CTM \citep{bianchi_pre-training_2021} models we trained by coherence (NPMI), diversity, and topic number. We trained the model on a dataset of 183 citation statements which are in-text passages from publications citing \citep{lau_machine_2014} retrieved from scite.ai \citep{nicholson_scite_2021}. This dataset was also used as the reference corpus for evaluating NPMI in this initial case study. We acknowledge that when building robust topic models a standard reference corpus should be used so results can be comparable in future works.

\section{Topic Representations}
\label{outputs}
Three randomly sampled topics are presented here so readers can inspect the quality of the topic labels as well as the original topic labels produced by CTM.

\begin{table*}[t!]
\centering
\begin{tabular}{p{0.9\linewidth}} \hline
Model: topic lists \\ \hline 
measure perplexity npmi probability paper correlation high log\_p ranking correspond \\ \hline
Model: ms2 \\ \hline
In conclusion, NPMI has a positive correlation with the human evaluation of topic coherence. \\ \hline
Model: multixscience \\ \hline
NPMI has been shown to correlate better with human evaluation of topic coherence than PMI cite cite cite. The NPMI measure was first introduced by Lau et. al cite and has been used extensively in the literature since cite. It is used to evaluate the coherence of the top @math words of a topic and to measure the difficulty of associating an individual topic with a single semantic concept, and consequently, evaluating topic models by their interpretability. It has also been used as a measure of the performance of automatic topic evaluation methods, and has shown to be more correlated with human performance than other traditional topic measures such as perplexity cite, co-occurrence cite cite, and held-out likelihood cite. In this paper, we focus on NPM I since it has been found to have a higher correlation with the human evaluation than other topic measures, such as PMI, and will be used as the evaluation measure in this paper. We refer the reader to cite cite for more details. \\ \hline
Model: multi-lexsum-tiny \\ \hline
NPMI and Topic Coherence are measures used to measure the semantic coherence of topics. \\ \hline
Model: multi-lexsum-long \\ \hline
For simplicity, we will refer to this measure as NPMI since it has been shown to have a higher correlation with the human evaluation of topic coherence than PMI (Lau et al, 2014). For more on this topic, see our related article. Introduction to Topic Coherence. In the literature, human-annotated topic models are usually measured using the Coherence and Normalized Pointwise Mutual Information (NPMI), which consider the probability that both terms in a particular term pair, where frequent terms will be downweighted with respect to those that occur less frequently, such as education from this example. However, in this paper, the authors adapted it to measure the difficulty of associating an individual topic with a single semantic concept, and consequently, evaluating topic models by their interpretability. The authors of this paper also used the predictive Probability and Predictive Probability to see the performance of the learning methods. The results of this study were shown to be significantly improved compared with the counterparts without PPMI, especially in terms of perplexity and topic-coherence and with low proportion of the training words. The performance measure was estimated using 2 measures to evaluate the methods: Log predictive probability (LPP) which considers the generalization of a model and Normalization pointwise mutual information (NPI) which evaluates the coherence in the most significant words (top words) of a topic. The coherence score was estimated through the point-wars mutual information. The Coherence Score was calculated using the standard linear predictive probability method and was estimated to correlate well with human judging. \\  \hline
\end{tabular}
\caption{
\label{topic_0}
Outputs for topic \#0
}
\end{table*}

\begin{table*}[t]
\centering
\begin{tabular}{p{0.9\linewidth}}\hline
Model: topic lists \\ \hline
automate measure mutual correlate röderal coherence draw recent pointwise topic \\ \hline
Model: ms2 \\ \hline
The meta- analysis showed that topic coherence was positively correlated with the interpretability of the topic models.\\ In contrast, the meta-analyses did not find a significant association between topic co-occurrence and interpretability. \\ \hline
Model: multixscience \\ \hline
Topic coherence is a measure of the interpretability of a topic model. It is based on the coherence of the words assigned to a topic and has been studied extensively in the context of topic modeling. Various measures have been proposed to measure topic coherence, such as the pointwise mutual information (PMI) between the topic words and the co-occurrence frequency of these words in the reference corpus, as well as the number of topics in the model. The PMI-based methods have been widely used in the evaluation of topic models (see for example cite cite and the references therein). However, these methods do not take into account the internal representation of the topic models. To the best of our knowledge, there is no prior work that evaluates the topic interpretability by measuring the PMI. However, there has been a large body of work on evaluating topic models by measuring their interpretability, including methods based on model perplexity, coherence, predictiveness cite, NPMI, topic diversity, and distributional semantics \\ \hline
Model: multi-lexsum-tiny \\ \hline
Topic model quality and interpretability are two different metrics used to measure the semantic interpretability of a topic. \\ \hline
Model: multi-lexsum-long \\ \hline
More specifically, Chang et al showed that models that fare better in predictive perplexity often have less interpretable topics, suggesting that evaluation should consider the internal representation of topic models and aim to quantify their interpretability. The idea soon gave rise to a new family of methods (Newman et al, 2010). Auto-Auto-NPMI that evaluate the semantic interpretability by measuring the number of chosen topics. These methods assume that topic coherence correlates with the coherence of the words assigned to that topic and thus quantify topic model quality. The resulting output does not explicitly provide an interpretation to human users, but remains a list of words with associated topic probabilities, which humans must interpret (as rated by human annotators) despite the observation that evaluation methods such as perplexity are often not correlated with human judgements of topic quality. According to the most recent work in topic design evaluation, indirect approach with word intrusion, and direct approach with observed coherence are two major approaches to measure the human interpretability in ML algorithms. The direct approach is to choose the model with the lowest perplexity on unseen data, which may generate topics that are hardly interpretable, and the direct approach uses the automated evaluation measure OC-Auto NPMI in Grieser et al. that evaluates the co-occurrence frequencies of terms within a reference corpus. Both methods have been developed to provide empirical evidence for the number chosen topics, because they are based on topic diversity; it is the percentage of unique words in the top 25 words of all topics, without mention of obstruction 5/10/20 topic words. The second method is to use coherence, which is to compute topic diversity by measuring topic diversity. The third method uses coherence to offer empirical evidence. \\ \hline
\end{tabular}
\caption{
\label{topic_4}
Outputs for topic \#4
}
\end{table*}

\begin{table*}[t!]
\centering
\begin{tabular}{p{0.9\linewidth}} \hline
Model: topic lists \\ \hline
compute word mutual coherence draw interpretability similarity assess pointwise human \\ \hline
Model: ms2 \\ \hline
Meta-analyses of topic coherence showed that the coherence of topics was in line with human evaluation of the interpretability of the topics. \\ \hline
Model: multixscience \\ \hline
Topic coherence is a quantitative measure of the interpretability of individual topics. It is the average pointwise mutual information of two words drawn randomly from the same document cite. The coherence between top words within a topic is estimated using the PMI between topic words cite cite cite. Various formulations have been proposed to compute topic coherence, including those based on the NPMI cite cite, PMI and its variations cite, the Normalised PMI cite, and the Point-wise Mutual Information (PMI) cite. Topic coherence scores judged by human annotators cite cite are used as a measure of topic interpretability. The most popular evaluation metrics are LPP cite, which measures the generalization of a topic model on unseen data, and NPMI cite, that measures the coherence of the topics. However, LPP is not the best measure for evaluating topic coherency. \\ \hline
Model: multi-lexsum-tiny \\ \hline 
Evaluation metrics: Log predictive probability (LPP) and topic interpretability \\ \hline
Model: multi-lexsum-long \\ \hline
Evaluation metrics: Log predictive probability (LPP) and Normalized pointwise mutual information (NPMI) are used. While LPP measures the generalization of a model on unseen data, NPMI examines the coherence and interpretability of the learned topics. For each topic t, Experiments show topic coherence (TC), which is in line with human evaluation of topic interpretability, and Experiments ShowTopic Coherence Experiments (TC) computed with the Coherence between a topic's most representative words (e.g., top 10 words) is inline with human eval of topic interpretationability. As the reference corpus for computing word occurrences, we use the English Wikipedia. As various formulations have been proposed to compute TC, we refer readers to Röder et al. (2015) for more concrete ways to see how the topic models interact with each other. To quantitatively measure the interpretability or the semantic quality of individual topics, we used the observed coherence measure from (Lau et al., 2014), which was adopted from psychology theory and showed better topic interpretation compared with other measures {[}1, 2{]}. In addition to the above measures, we looked for the observed relationship between the topic and human interpretation of topic models. The observed correlation between the top N words within a topic and its coherence between the bottom 10 words was inline with the human evaluation in evaluations 2-5 8. It is a preferred method for such tasks (Aletras and Stevenson, 2013;Newman and al, 2010a) as it is unaffected by variability in the range for each dataset. \\ \hline
\end{tabular}
\caption{
\label{topic_2}
Outputs for topic \#2
}
\end{table*}

\end{document}